\documentclass[a4paper]{svproc}
\usepackage{makeidx}
\makeindex
\usepackage[table]{xcolor}
\usepackage{graphicx,amsmath,amssymb,epstopdf,mathtools,multirow,booktabs}
\usepackage{subcaption}	
\usepackage{threeparttable, tablefootnote}
\usepackage{arydshln}
\usepackage{pdfrender}
\usepackage{cite}
\usepackage[hidelinks]{hyperref}
\usepackage{paralist}
\usepackage{tikz}
\usepackage{algorithmic}
\usepackage{algorithm}
\floatstyle{ruled}

\newcommand{\trsp}{{\!\scriptscriptstyle\top}}

\let\vec\boldvec

\author{Shaokang Wu, Yijin Wang \and Yanlong Huang}
\authorrunning{Shaokang Wu et al.}

\institute{School of Computing, University of Leeds, Leeds LS29JT, UK. \\
\email{ \{scswu, ml20y3w, y.l.huang\}@leeds.ac.uk }}
\title{One-Shot Robust Imitation Learning for Long-Horizon Visuomotor Tasks from Unsegmented Demonstrations}
\titlerunning{Meta-Imitation Learning with Adaptive Dynamical Primitives}

\begin{document}
\maketitle

\begin{abstract}
In contrast to single-skill tasks, long-horizon tasks play a crucial role in our daily life, e.g., a pouring task requires a proper concatenation of reaching, grasping and pouring subtasks.  As an efficient solution for transferring human skills to robots, imitation learning has achieved great progress over the last two decades. 
However, when learning long-horizon visuomotor skills,
imitation learning often demands 
a large amount of semantically segmented demonstrations.
Moreover, the performance of imitation learning could be susceptible to external perturbation and visual occlusion. 
In this paper, we exploit dynamical movement primitives and meta-learning to provide a new framework for imitation learning, called
\textbf{M}eta-\textbf{I}mitation \textbf{L}earning with \textbf{A}daptive Dynamical Primitives (MiLa).  
MiLa allows for learning unsegmented long-horizon demonstrations and adapting to unseen tasks with a single demonstration. MiLa can also 
resist external disturbances and visual occlusion during task execution. 
Real-world robotic experiments demonstrate the superiority of MiLa, irrespective of visual occlusion and random perturbations on robots.
\end{abstract}


\section{Introduction}  \label{sec:intro}
Learning long-horizon visuomotor tasks is challenging due to dynamical visual observations and long-horizon decision-making processes. In the context of imitation learning, many methods have been developed towards solving long-horizon tasks \cite{xu2023xskill, RT-1, RT-Sketch,daml_long_horizion_tasks, mimicplay}. However, these approaches may show limited generalization to new objects \cite{xu2023xskill}, require an impractical number of demonstrations \cite{RT-1, RT-Sketch}, or become inefficient when learning from unsegmented demonstrations \cite{daml_long_horizion_tasks}. In addition, imitation learning can be easily susceptible to external perturbations \cite{mimicplay} (e.g., the robot's proprioceptive states dramatically change as a consequence of sudden interventions from human users) and visual occlusion, which further impedes its deployment in real robotics tasks.

One promising avenue to achieve perturbation resilience is to encode skills via an autonomous dynamical system (DS) \cite{DMP, SEDS, ijspeert2002movement, saveriano2023dynamic}. Unlike end-to-end mapping using deep neural networks, DS can handle out-of-distribution states and resist disturbances, where the convergence property of DS is theoretically guaranteed. However, DS usually requires predefined task parameters (e.g., the 3D location of an object in a reaching task), limiting its scalability to high-dimensional visuomotor tasks where only visual observations are available.

In this paper, we aim to endow robots with the capability of rapidly acquiring new long-horizon visuomotor tasks while ensuring robustness throughout task execution. To do so, we propose a novel approach called \textbf{M}eta-\textbf{I}mitation \textbf{L}earning with \textbf{A}daptive dynamical primitives (MiLa), which leverages meta-imitation learning for one-shot learning and dynamical movement primitives (DMP) for robustness and smoothness in task execution. Specifically, MiLa enables the learning of unsegmented long-horizon demonstrations, without additional semantic parsing or phase prediction for demonstration segmentation \cite{daml_long_horizion_tasks}.

MiLa is predicated on the assumption that long-horizon tasks are composed of elementary subtasks. We build a skill repertoire consisting of movement primitives and encode them using a well-known dynamical system approach DMP, where only a single demonstration is needed for each primitive. On top of that, we learn a high-level policy via meta-imitation learning to predict the corresponding task parameters from visual observations for each subtask. Within each subtask, the predicted task parameters are fixed and the entire sequence of robot actions is generated using the appropriate DMP under the predicted task parameters. Since each motion primitive represents a type of motion pattern (e.g., reaching, placing, or pushing), robot actions throughout the same subtask are expected to be more consistent and legible, as opposed to the continuous prediction of robot actions as per the current visual observations \cite{daml_long_horizion_tasks,mimicplay}.

The contribution of this paper is a robust meta-imitation learning framework that is capable of\begin{enumerate}
    \item[(\emph{i})] learning from unsegmented long-horizon visuomotor demonstrations,
    \item[(\emph{ii})] adapting to new tasks with only one-shot demonstration,
    \item[(\emph{iii})] resisting external perturbations on robots and visual observations. 
\end{enumerate}

\section{Related Work} \label{sec:related}
\subsection{Imitation Learning of Movement Trajectories}

Many imitation learning algorithms with a focus on motion planning have been proposed and various successful applications have been reported \cite{ijspeert2002movement, ProMP, KMP}. For instance, DMP learns the motion pattern of a single demonstration using a spring-damper system, wherein the equilibrium of the system corresponds to the desired target of the robot's motion. Probabilistic movement primitives (ProMP) \cite{ProMP} and kernelized movement primitives (KMP) \cite{KMP} respectively employ basis and kernel functions to capture the probabilistic characteristics of multiple demonstrations. These approaches exhibit high sample efficiency, allowing for skill learning from just one or a few demonstrations. However, this type of approach targets the learning of demonstrations associated with time input or multiple-dimensional inputs and becomes inappropriate when dealing with high-dimensional visual inputs (i.e., images).

To connect imitation learning for motion planning with image inputs, there are some works on learning a mapping from images to DMP parameters (e.g., basis function weights, motion target and duration) \cite{NDPs, gams2018deep, H-NDPs}. Once these parameters are obtained, smooth trajectories can be naturally generated via DMP. However, these approaches are restricted to the learning of single-skill tasks.

\subsection{Imitation Learning of Long-Horizon Tasks}

Many recent works addressed long-horizon visuomotor tasks via imitation learning. In \cite{GTI, shankar2019discovering, nasiriany2022learning, sharma2019third}, hierarchical imitation learning was studied, where a high-level policy was used to make `plans' for compound tasks and a low-level controller was designed to execute these `plans' (e.g., latent variables or sub-goals). In \cite{shin2023one, duan2017one}, one-shot imitation learning was investigated, aiming to learn a new task from a single demonstration in the form of a complete or partial trajectory and video. Notably, meta-imitation learning has earned a relevant place due to its reliable performance \cite{mil,li2021meta, daml}. The objective of meta-imitation learning is to exploit shared structures among the tasks sampled from the same distribution and search for an optimal policy capable of adapting quickly to new tasks. Most works in this line only consider learning a single skill, without tapping the learning of long-horizon skills.

A work on meta-learning that is closely related to ours is \cite{daml_long_horizion_tasks}, which tackles long-horizon tasks by segmenting them into subtasks and subsequently performing meta-imitation learning on subtasks. However, this approach requires primitive-level demonstrations to train an additional motion phase predictor. The predictor is used for segmenting demonstrations into subtasks, acting as an indispensable step during meta-training. In fact, collecting unsegmented long-horizon demonstrations is more straightforward and raw demonstrations are more easily accessible. In this paper, we propose to meta-train a policy on unsegmented visuomotor demonstrations, without any specific treatment on demonstration parsing or segmentation. Besides, we leverage the dynamical feature of DMP to ensure that our framework is robust to external perturbations (e.g., from visual inputs and the robot's proprioceptive states), leading to another advantage against existing imitation learning methods \cite{mil,li2021meta, daml,mimicplay, xu2023xskill}.

\section{Preliminaries} \label{sec:pre}
\subsection{Dynamical Movement Primitives} \label{subsec:dmp}

Suppose we have access to a demonstration of time-length $N$, i.e., $\{t_n, \vec{\xi}_n,\dot{\vec{\xi}}_n,\ddot{\vec{\xi}}_n\}_{n=1}^{N}$. Here, $\vec{\xi}_n \in \mathbb{R}^\mathcal{O}$ represents $\mathcal{O}$-dimensional position (or joint angles) at the $n$--th time step, while $\dot{\vec{\xi}}_n$ and $\ddot{\vec{\xi}}_n$ respectively denote the corresponding velocity and acceleration. DMP encodes the demonstration using a second-order dynamical model:
\begin{equation}
\begin{aligned}
\label{equ:dmp-phase}
&\tau \dot{s} = - \alpha s,
\end{aligned}
\end{equation}
\begin{equation}
\begin{aligned}
\label{equ:dmp-traj}
&\tau^2 \Ddot{\vec{\xi}} = \vec{K}_p (\vec{g} - \vec{\xi}) - \tau \vec{K}_v \dot{\vec{\xi}} + s (\vec{g} - \vec{\xi}_0) \odot \boldsymbol{f}_{\vec{w}}(s),
\end{aligned}
\end{equation}
\begin{equation}
\begin{aligned}
\label{equ:dmp-force}
&\boldsymbol{f}_{w}(s) = \vec{W} \bigg[ \frac{\varphi_1(s)}{\sum_{h=1}^H \varphi_h(s)}  \  \frac{\varphi_2(s)}{\sum_{h=1}^H \varphi_h(s)}  \  \cdots  \  \frac{\varphi_H(s)}{\sum_{h=1}^H \varphi_h(s)} \bigg]^\trsp.
\end{aligned}
\end{equation}
Equation (\ref{equ:dmp-phase}) is utilized to convert time into the phase variable $s$, thereby eliminating explicit time dependence. Here, $\tau$ denotes the motion duration, and $\alpha$ signifies the decay factor. In (\ref{equ:dmp-traj}), $\vec{K}_p$ and $\vec{K}_v$ denote the user-specified stiffness and damping matrices, respectively. $\vec{g}$ and $\vec{\xi}_0$ represent the goal (end-point) and start-point of a trajectory. The symbol $\odot$ stands for the element-wise product. $\boldsymbol{f}_{w}(s)$ represents the forcing term, typically expressed as a linear combination of pre-defined Gaussian basis functions (see (\ref{equ:dmp-force})). $\vec{W} \in \mathbb{R}^{\mathcal{O} \times H} $ means learnable parameters corresponding to the motion pattern of the demonstration.

\subsection{Model-Agnostic Meta-learning}  \label{subsec:maml}

\newcommand{\task}{\mathcal{T}}
\newcommand{\loss}{\mathcal{L}}
\newcommand{\inp}{\mathbf{x}}
\newcommand{\learner}{f}
\newcommand{\lossi}{\loss_{\task_i}}

\newcommand{\data}{\mathcal{D}}
\newcommand{\humandata}{\data^h}
\newcommand{\humandemo}{\mathbf{d}^h}
\newcommand{\demo}{\mathbf{d}}
\newcommand{\robotdata}{\data^r}
\newcommand{\robotdemo}{\mathbf{d}^r}
\newcommand{\learnedloss}{\loss_\psi}
\newcommand{\bcloss}{\loss_\text{BC}}

Model-agnostic meta-learning (MAML) is a meta-learning algorithm proposed to rapidly learn new tasks using a small number of data \cite{maml}. It operates under the assumption that a shared structure exists among meta-training and meta-test tasks (i.e., all tasks are drawn from the same task distribution). Consider imitation learning using MAML with a policy $\pi_\theta$ parameterized by $\vec{\theta}$ and gauged by behaviour cloning (BC) loss function $\bcloss$. During meta-training, MAML randomly selects a meta-training task $\task$ from the task distribution $p(\task)$ and partitions demonstrations from the task $\task$ into training dataset $\data^\text{tr}_\task$ and validation dataset $\data^\text{val}_\task$. MAML then optimizes the policy parameters $\vec{\theta}$ such that one (or a few) gradient update on $\data^\text{tr}_\task$ leads to favourable performance on $\data^\text{val}_\task$. The objective function of MAML is formulated as
\begin{equation}
\begin{aligned}
    &\min_{\vec{\theta}}  \sum_{\task \sim p(\task)} \bcloss(\vec{\theta}-\alpha \nabla_{\vec{\theta}} \bcloss(\vec{\theta}, \data^\text{tr}_\task), \data^\text{val}_\task) \\
    &= \min_{\vec{\theta}} \sum_{\task \sim p(\task)} \bcloss(\vec{\phi}_\task, \data^\text{val}_\task),
    \vspace{-0.1cm}
\end{aligned}
\end{equation}
where $\alpha>0$ refers to the step size of gradient descent, and $\vec{\phi}_\task$ corresponds to the updated parameters after learning on $\data^\text{tr}_\task$. During the meta-testing (i.e., model inference) phase, the policy adapts to a new task $\data_{\task_\text{test}}$ by utilizing the updated  parameters $\vec{\phi}_{\task_\text{test}} = \vec{\theta} - \alpha \nabla_\theta \bcloss(\vec{\theta}, \data_{\task_\text{test}}).
$

\section{Meta-Imitation Learning with Adaptive Dynamical Primitives} \label{sec:method}

We assume that we have access to a dataset $\{\vec{d}_h\}_{h=1}^{H}$ across $K$ tasks $\{\task_k\}_{k=1}^K$, with $\vec{d}_h$ represents the $h$-th
unsegmented, long-horizon visuomotor demonstration. Each demonstration $\vec{d}_h=\{\vec{v}_h, \vec{\kappa}_h\}$ consists of a sequence of visual images  $\vec{v}_h = \{ \vec{o}_t \}_{t=1}^{T(h)}$, as well as the corresponding robot trajectory $\vec{\kappa}_h = \{\vec{\xi}_t\}_{t=1}^{T(h)}$. Here, $T(h)$ denotes the time length, $\vec{o}_t$ and $\vec{\xi}_t$ represent image observation and robot action (e.g., end-effector position or joint angles), respectively. We assume all demonstrations can be decomposed into $C$ different subtasks and the order of the subtasks is known.

To learn from unsegmented demonstration, we first establish a repertoire of motion primitives and use DMP to encode each primitive. Given the elementary motion primitives, we propose to learn a high-level policy, referred to as MiLa, for predicting task parameters for motion primitives and composing primitives.

\subsection{Skill repertoire} \label{subsec:skill}

The objective of constructing a skill repertoire is to provide a set of reliable primitives for a high-level policy. By reusing and composing motion primitives, complex and long-horizon tasks can be accomplished. For each type of skill (e.g., reaching, placing, and pushing), we collect one demonstration of robot trajectory via kinesthetic teaching. By feeding the demonstration into the DMP model in (\ref{equ:dmp-force}), the motion pattern that underlies the demonstration can be determined. Repeating the same procedure for all primitives, we can obtain a skill repertoire comprising different DMPs, denoted by $\{\rho^c\}_{c=1}^C$.

Given a DMP $\rho^c$, we can use it to generate an adapted trajectory for a new task, where we only need to specify starting point $\vec{\xi}_0$, desired target $\vec{g}$, and motion duration $\tau$. The adapted trajectory is given by
\begin{equation}
\begin{aligned}
\hat{\vec{\kappa}}^c(t) &= \rho^c(\vec{\xi}_0, \vec{g}, \tau, t),
\end{aligned}
\end{equation}
where 
$\hat{\vec{\kappa}}^c(t) \in \mathbb{R}^\mathcal{O}$ denotes the planned robot actions at time $t$ via $\rho^c$. Therefore, DMP builds a connection between task parameters (i.e., $\{\vec{\xi}_0, \vec{g}, \tau \}$) and robot trajectories. Specifically, the adapted trajectory via DMP maintains the motion style extracted from the demonstration, which is expected to be legible and predictable by human users.

\subsection{Meta-Imitation Learning with Adaptive Dynamical Primitives}
\label{subsec:planner}

\begin{figure*}[bt]
    \centering
\includegraphics[width=0.95\linewidth]{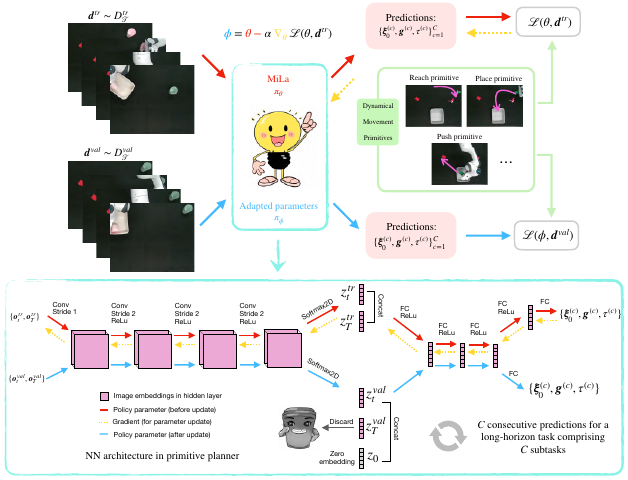}
    \caption{An overview of the MiLa framework. After learning from $\vec{d}^{\text{tr}}$ (indicated by yellow arrows), MiLa acquires the capability to adapt to new tasks $\vec{d}^{\text{val}}$ (see blue arrows). Instead of predicting robot actions as per visual inputs at each time step, MiLa predicts task parameters for each subtask and a set of dynamical movement primitives are employed to generate robot trajectories across different subtasks.}
    \label{fig:MiLa-training}
\end{figure*}

Instead of predicting robot actions according to visual inputs at each time step, we propose to predict task parameters for each subtask instead. Formally, given an initial image observation $\vec{o}_1$ and the goal image $\vec{o}_T$ (i.e., the last observation) of a demonstration, we propose to learn the policy $\pi_\theta (\vec{\xi}_0^{(1)}, \vec{g}^{(1)}, \tau^{(1)} | \vec{o}_1, \vec{o}_T)$ that predicts the start-point, end-point and motion duration of the first subtask. 
The predicted task parameters $(\vec{\xi}_0^{(1)}, \vec{g}^{(1)}, \tau^{(1)})$ will be passed onto the motion primitive $\rho^1$, yielding a robotic trajectory $\hat{\vec{\kappa}}^1(t)=\rho^1(\vec{\xi}_0^{(1)}, \vec{g}^{(1)}, \tau^{(1)},t)$. Similarly, new task parameters $(\vec{\xi}_0^{(2)}, \vec{g}^{(2)}, \tau^{(2)})$ for the second subtask are predicted using observations at $\vec{o}_{1+\tau^{(1)} / \delta}$ together with the goal image $\vec{o}_T$, and a new robot trajectory is planned as $\hat{\vec{\kappa}}^2(t)=\rho^2(\vec{\xi}_0^{(2)}, \vec{g}^{(2)}, \tau^{(2)},t)$. Here, $\delta$ denotes the time interval between two consecutive robot actions. By repeating the prediction procedure, we can predict $C$ groups of task parameters and generate $C$ trajectories for the robot. Finally, we can obtain the entire robot trajectory, i.e., $\hat{\vec{\kappa}} = \hat{\vec{\kappa}}^1 \oplus \hat{\vec{\kappa}}^2 \cdots \oplus \hat{\vec{\kappa}}^C$, where $\hat{\vec{\kappa}}^c$ represent the trajectory in subtask $c$ and $\oplus$ denotes the concatenation of trajectories.

Before we measure the difference between the predicted robot trajectory $\hat{\vec{\kappa}}$ and the demonstrated robot trajectory $\vec{\kappa}^h = \{\vec{\xi}_t\}_{t=1}^{T(h)}$, we exploit the covariance weighted loss function. As pointed out in numerous works on probabilistic imitation learning \cite{ProMP, huang2018generalized, TP-GMM}, different demonstrations could be collected even for the same task and such variability is naturally reflected by the variance of demonstrations. In addition to the demonstrations collected for training DMPs in Section~\ref{subsec:skill}, we collect additional (approximately 5--6) demonstrations to model the intrinsic variability for each type of skill. We use Gaussian mixture model to learn the joint distribution $\mathcal{P}(t, \vec{\xi})$ and adopt Gaussian mixture regression \cite{GMR} to compute the covariance function $\vec{\Sigma}^c(t)= \mathbb{D}(\vec{\xi}(t) | t)$. Now, we formulate the covariance weighted loss function between the predicted and demonstrated long-horizon robot trajectories as
\begin{equation}
\mathcal{L}_{cov} (\vec{\theta}, \vec{d}_h)= \frac{1}{T(h)} \sum_{c=1}^C \sum_{t=1}^{\tau^{(c)}} \gamma_c \big( \vec{\kappa}^h(t_c+t) - \hat{\vec{\kappa}}^c(t) \big)^\trsp \big(\vec{\Sigma}^c(t) \big)^{-1} \big( \vec{\kappa}^h(t_c+t) - \hat{\vec{\kappa}}^c(t) \big),
\label{equ:mila:loss}
\end{equation}
where $t_c=\sum_{i=1}^{c-1}\tau^{(i)}$ ($t_c=0$ if $c=1$), $\gamma_c$ represents an adjustable weight parameter for each subtask. The entire loss function for meta-training becomes
\begin{equation}
\begin{aligned}
J(\vec{\theta})=\sum_{\task \sim p(\task)} \sum_{ (\vec{d}^\text{tr}, \vec{d}^\text{val}) \in \data_\task } \mathcal{L}_{cov}  (\vec{\phi}_\task, \vec{d}^\text{val}) \\
\mathrm{with}\quad \vec{\phi}_\task = \vec{\theta} - \alpha \nabla_{\vec{\theta}} \mathcal{L}_{cov} (\vec{\theta}, \vec{d}^\text{tr}),
\end{aligned}
\label{equ:mila:update}
\end{equation}

With the loss function in (\ref{equ:mila:loss}), we can compute the training error without segmenting demonstrations beforehand, which largely facilitates meta-training on unsegmented demonstrations. Note that $\mathcal{L}_{cov}$ achieves its minimum value only when the predicted subtasks (i.e., task parameters) are accurate. Therefore, minimizing the loss function in (\ref{equ:mila:loss}) drives the policy towards learning all subtasks in each long-horizon task precisely. An overview of the proposed approach MiLa is depicted in Fig.~\ref{fig:MiLa-training}. To comply with meta-testing where the goal image is not provided after the one-shot demonstration, we sample a random image as the goal image $\vec{o}_T^{val}$ and replace the embedding of the goal image with a vector $\vec{z}_0$ during the meta-training phase. The entire procedure of MiLa is summarized in Algorithm~\ref{alg:mila}.

\begin{algorithm}[t]
\caption{Meta-Imitation Learning with Adaptive Dynamical Primitives}
\label{alg:mila}
\begin{algorithmic}[1]
\REQUIRE $\alpha$: step size for gradient descent
\REQUIRE skill repertoire $\rho^c$ and covariance function $\vec{\Sigma}^c(t)$, $c=1,2,\ldots,C$
\STATE randomly initialize $\vec{\theta}$
\WHILE{meta-training}
\STATE Sample a task $\task \sim p(\task)$
\STATE Random partition $\data_\task$ as $(\data^\text{tr}_{\task}, \data^\text{val}_{\task})$
\STATE Sample one demonstration $\vec{d}^\text{tr} \sim \data_\task^{\text{tr}}$
\STATE Predict $C$ groups of task parameters along $\vec{d}^\text{tr}$
\STATE Generate robot trajectories for all subtasks with task parameters from line 6 
\STATE Compute adapted parameters $\vec{\phi}_\task = \vec{\theta} - \alpha \nabla_{\vec{\theta}} \mathcal{L}(\vec{\theta}, \vec{d}^\text{tr})$ via (\ref{equ:mila:update})
\STATE Sample one demonstration $\vec{d}^{\text{val}} \sim \data_\task^{\text{val}}$
\STATE Predict $C$ groups of task parameters along $\vec{d}^\text{val}$
\STATE Generate robot trajectories for all subtasks with task parameters from line 10
\STATE Update parameters $\vec{\theta}$ via minimizing (\ref{equ:mila:update})
\ENDWHILE
\RETURN $\theta$
\end{algorithmic}
\end{algorithm}

MiLa predicts task parameters by `casting a glance' at the environment, it makes predictions solely from image observations at the beginning of each subtask, enabling it to resist visual occlusion over the course of the execution of the subtask. Furthermore, MiLa ensures that the robot convergences to the predicted target point owing to the dynamical feature offered by DMP \cite{DMP}, even when the robot encounters dramatic perturbations. Last, MiLa provides smooth trajectories for robots since the action trajectory for each subtask is planned as a whole by DMP.

\begin{figure*}[bt]
    \centering
\includegraphics[width=0.8\linewidth]{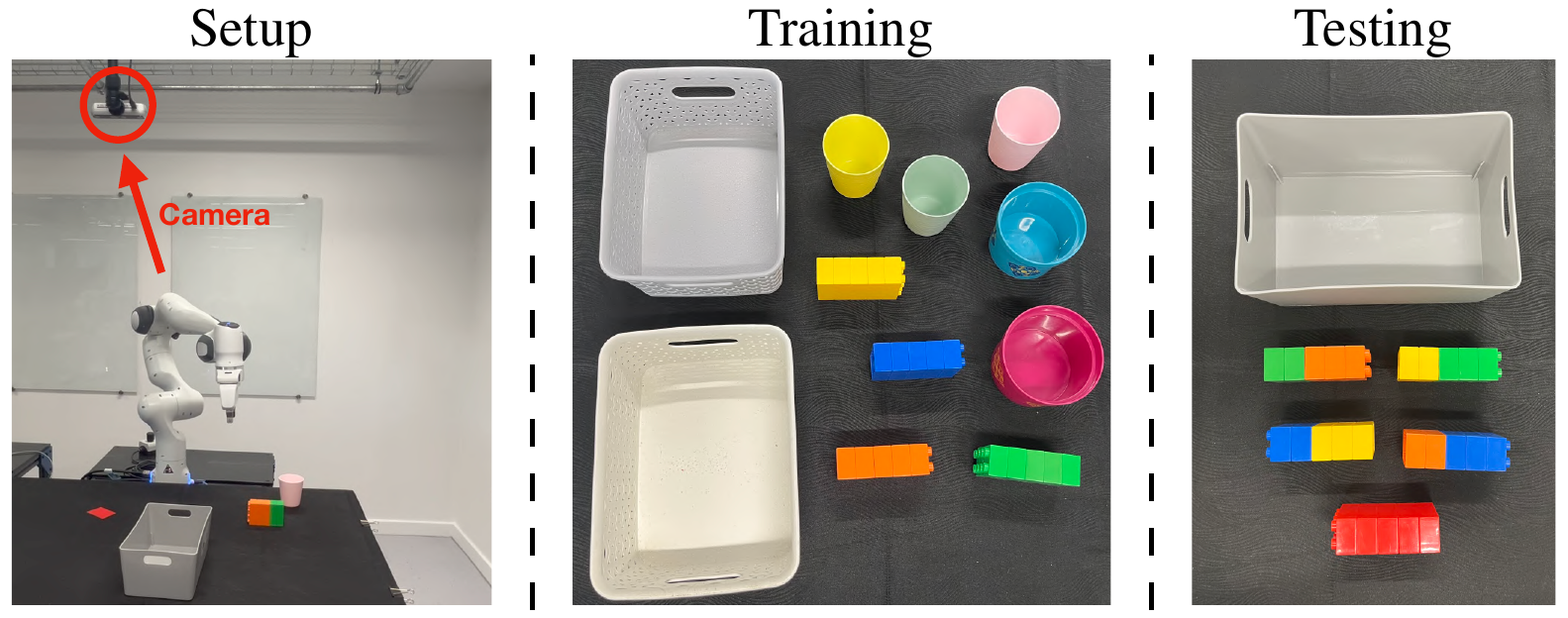}
    \caption{The experimental setup as well as objects for training and testing in long-horizon tasks.}
    \label{fig:objects}
\end{figure*}

\begin{figure*}[bt]
    \centering
\includegraphics[width=0.95\linewidth]{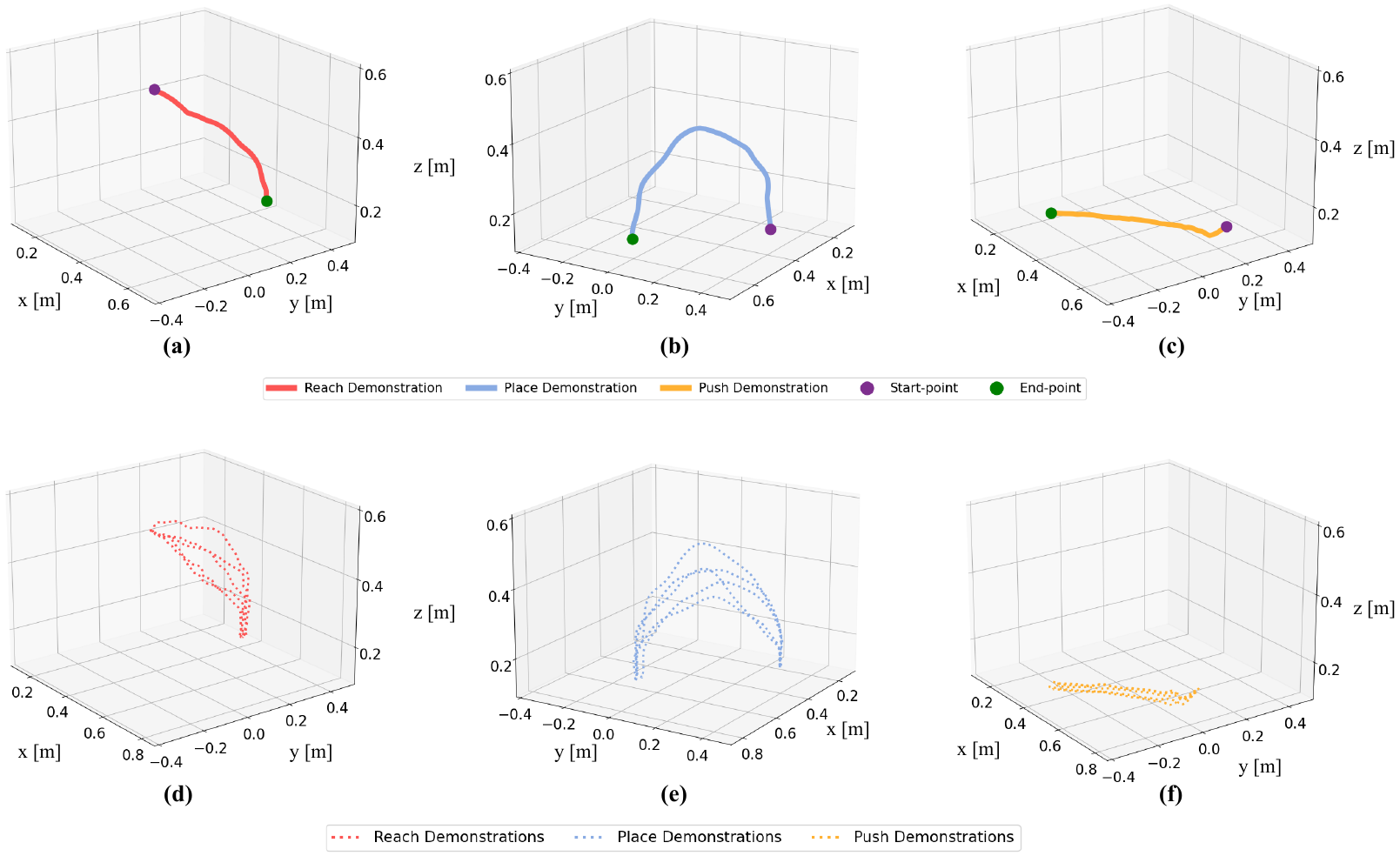}
    \caption{Demonstrations utilized to establish the skill repertoire $\{\rho^c\}_{c=1}^3$. Dynamical motion primitives for the reaching, placing, and pushing skills are learned from demonstrations depicted in (\emph{a})--(\emph{c}), respectively. Additionally, we collect 5 demonstrations for each skill to model the intrinsic skill variability, as shown in (\emph{d})--(\emph{f}).}
    \label{fig:skills-repertoire}
\end{figure*}

\section{Experiments} \label{sec:exp}

In this section, we aim to answer the following questions: 
\begin{itemize}
    \item[(1)] Is MiLa competitive with state-of-the-art baselines in long-horizon tasks, including MAML and goal-conditioned BC?
    \item[(2)] Can MiLa effectively resist visual occlusion and external disturbances during real-world task execution?
\end{itemize}

We consider the long-horizon visuomotor task that requires the robotic arm to reach and grasp a target object, place it into a basket, and finally push the basket to a desired location (i.e., the red squared marker). We collected 1,260 demonstrations as the training set using a 7-DoF Franka Emika Panda robotic arm and a top-view Intel RealSense D455 camera. The experimental setup is shown in Fig.~\ref{fig:objects} (left plot). The middle plot of Fig.~\ref{fig:objects} depicts the objects used for demonstration collection. All demonstrations, including videos and corresponding trajectories, were collected at the frequency of 30 Hz, where each demonstration lasts approximately 20 seconds. The demonstrations used to establish the skill repertoire and estimate the covariance function are depicted in Fig.~\ref{fig:skills-repertoire}.

We compare MiLa's performance against state-of-the-art approaches:
\begin{itemize}
    \item[$\bullet$] \textbf{Model-Agnostic Meta-Learning (MAML)}: a meta-imitation learning policy follows the implementation of \cite{daml_long_horizion_tasks, daml, maml, mil}, which requires the segmentation of long-horizon tasks into subtasks and then computes adapted parameters on the subtasks.
    \item[$\bullet$] \textbf{Goal-conditioned Behaviour Cloning (GCBC)}: a goal-image conditioned policy that takes as input a real-time image and the robot's current state, along with a final image of the subtask \cite{lynch2020learning}.
\end{itemize}
As an ablation study, we evaluate MiLa without using the covariance-weighted loss function, i.e., 
\begin{itemize}
    \item[$\bullet$] \textbf{MiLa-NoWeight}: setting $\vec{\Sigma}^c(t)$ as an identity matrix in (\ref{equ:mila:loss}).
 \end{itemize}

To ensure a fair comparison, all methods use a similar network architecture as illustrated in Fig.~\ref{fig:MiLa-training} and each method is evaluated with its optimal hyperparameters. As MAML requires the segmentation of long-horizon tasks into distinct subtasks, we train an individual model for each subtask. Similarly, GCBC also involves training an individual model for each subtask. In contrast, for MiLa and MiLa-NoWeight we train a single unified policy to directly learn the unsegmented, long-horizon demonstrations.

To assess the generalization capability in new settings (i.e., held-out objects, see the testing objects in Fig.~\ref{fig:objects}), we provide one-shot demonstration for MiLa, MiLa-NoWeight, and MAML to adapt their policy parameters. For GCBC, we collect goal images for each subtask separately per evaluation.

\subsection{Evaluations on reaching-placing-pushing tasks} \label{subsec:no-perturb}

\begin{figure*}[bt]
    \centering
\includegraphics[width=0.99\linewidth]{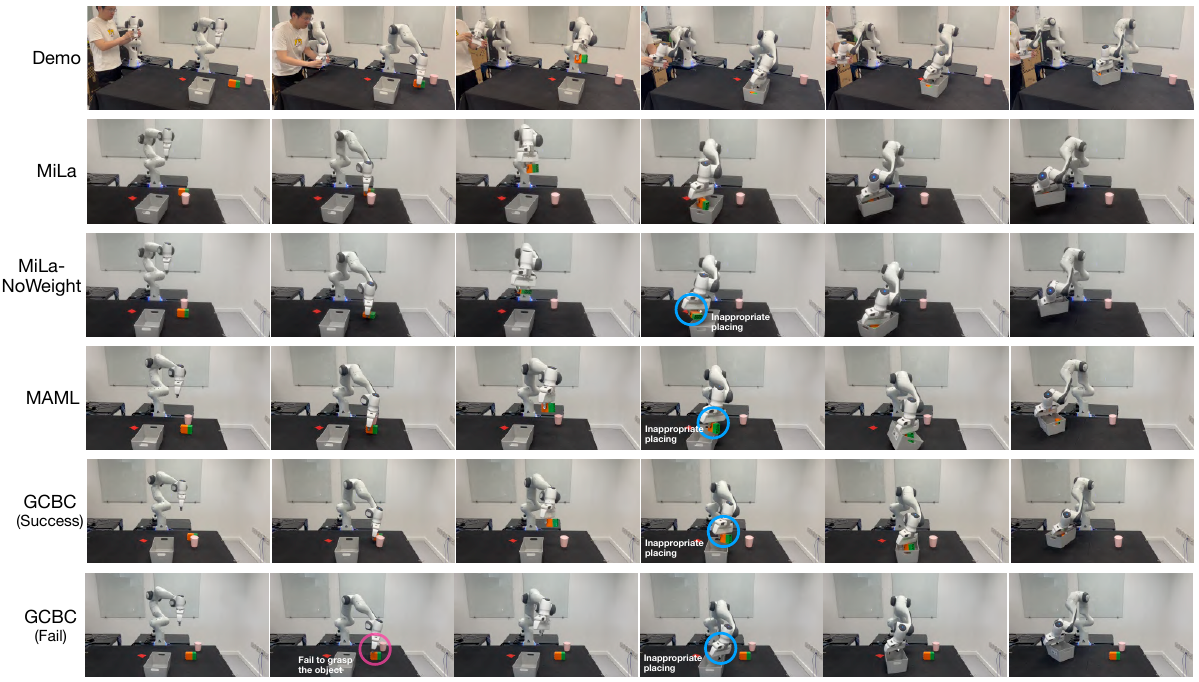}
    \caption{Snapshot of the long-horizon task evaluations. 
    \emph{First row} shows the kinesthetic teaching of the reaching-placing-pushing task. \emph{Second and third rows} correspond to the evaluations of MiLa and MiLa-NoWeight, respectively. \emph{Fourth row} illustrates an evaluation of MAML. \emph{Fifth and sixth rows} present the success and fail cases using GCBC, respectively.}
    \label{fig:snapshot-no-disturb}
\end{figure*}

\begin{table}[bt]
	\vspace{0.55cm}
      \centering
      \caption{Success rates of different methods.}
      \resizebox{0.99\columnwidth}{!}
      {
		\begin{tabular}{lccc}
			\toprule
			\textbf{Method} & Success \scriptsize{(misplacement)} \quad & Success \scriptsize{(proper placing)} \quad & Overall Success Rate  \\
                \midrule
                GCBC &  $25 \% $ & $20 \%$ & $45 \%$ \\
                MAML &  $40 \%$ & $5 \%$ & $45 \% $ \\
                MiLa-NoWeight & $15 \%$ & $55 \%$ & $70 \%$ \\
                MiLa & $\boldsymbol{0\%}$ & $\boldsymbol{70\%} $ & $\boldsymbol{70\%}$ \\
			\bottomrule
       \end{tabular}
    }
    \label{table:success-rate}
\end{table}

An illustration of the long-horizon task is provided in the first row of Fig~\ref{fig:snapshot-no-disturb}. We carry out 5 groups of evaluations on the held-out objects (see the right plot in Fig.~\ref{fig:objects}) and each group includes 4 trails by altering the locations of the target object and the basket. In total, we have 20 trials to evaluate each method.

The results in the second column of Table~\ref{table:success-rate} indicate that GCBC and MAML frequently place the grasped objects in the wrong places 
although the entire task is completed successfully due to the move and rotation of the basket, see examples in the fourth and fifth rows in Fig.~\ref{fig:snapshot-no-disturb}, where the grasped object collides with the basket. A smaller number of misplacements are also observed in MiLa-NoWeight, as displayed in the third row of Fig.~\ref{fig:snapshot-no-disturb}. In comparison, MiLa consistently places objects into the basket appropriately, see an example in the second row of Fig.~\ref{fig:snapshot-no-disturb}. This finding highlights the importance of using the covariance-weighted loss function to mitigate the effect of large skill variability in the placing task.

The fourth column of Table~\ref{table:success-rate} (i.e., the sum of the second and third columns) also shows that MiLa and MiLa-NoWeight achieve higher success rates than MAML and GCBC. We suggest that the low success rate in GCBC may be attributed to visual occlusion. The robot could obscure the object when approaching it since the camera is mounted over the object (see the left plot in Fig.~\ref{fig:objects}). In comparison, MiLa requires only a single image to perform each subtask, effectively mitigating issues related to visual occlusion. Finally, we find that many failure cases across these four approaches are attributable to the reaching subtask. Although the robot's gripper either touches or nearly touches parts of the objects, it still fails to grasp them. This issue has also been observed in \cite{daml_long_horizion_tasks}. Considering that MiLa-NoWeight is an ablation study of MiLa and its performance is inferior to MiLa, we only use MiLa for comparison with MAML and GCBC in the following evaluations.

\subsection{Evaluations in the presence of visual occlusion} \label{subsec:occlusion}

Now, we evaluate the performance of different methods by considering visual occlusion during task execution. In our evaluations,  visual occlusion is caused by the user's hand moving in front of the camera. The extreme case corresponds to the full occlusion of the camera's view. 

\begin{figure*}[bt]
    \centering
\includegraphics[width=0.99\linewidth]{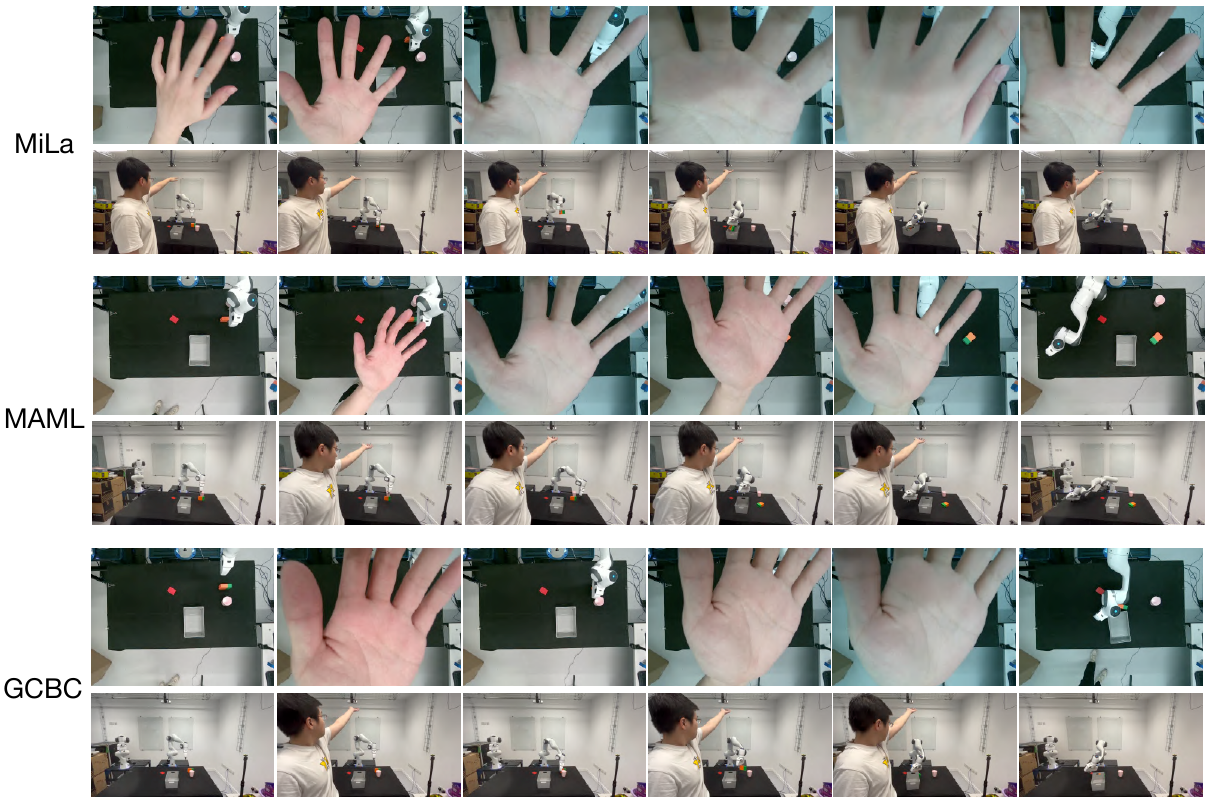}
    \caption{Snapshot of the long-horizon task in the presence of visual occlusion.  \emph{First and second rows} correspond to evaluations using MiLa, with the depth camera's perspective and the user's view, respectively. Similarly, \emph{third and fourth rows} correspond to evaluations using MAML. \emph{Fifth and sixth rows} are evaluations with GCBC.}
    \label{fig:snapshot-visual-disturb}
\end{figure*}

The snapshots illustrating the evaluation of MiLa under visual occlusion 
are presented in the first and second rows of Fig.~\ref{fig:snapshot-visual-disturb}. The first row captures the robot's perspective, while the second row provides the human user's view. Despite significant occlusion by the user's hand, MiLa successfully executes the long-horizon tasks. This success is achieved because MiLa processes only the first frame at the beginning of each subtask, thereby enabling it to disregard any subsequent frames affected by occlusion. In contrast, MAML and GCBC predict actions in a per-timestep or per-frame fashion and visual occlusion can adversely affect their predictions due to the introduction of out-of-distribution images, as illustrated in the third to sixth rows in Fig.~\ref{fig:snapshot-visual-disturb}, where both MAML and GCBC are unable to execute the placing task properly. Additionally, since MiLa operates at the trajectory level rather than on a per-timestep basis, this treatment provides smoother trajectories for the robot, compared to those generetated by MAML and GCBC.

\subsection{Evaluations in the presence of external pertubations} \label{subsec:perturbation}

\begin{figure*}[bt]
    \centering
\includegraphics[width=0.99\linewidth]{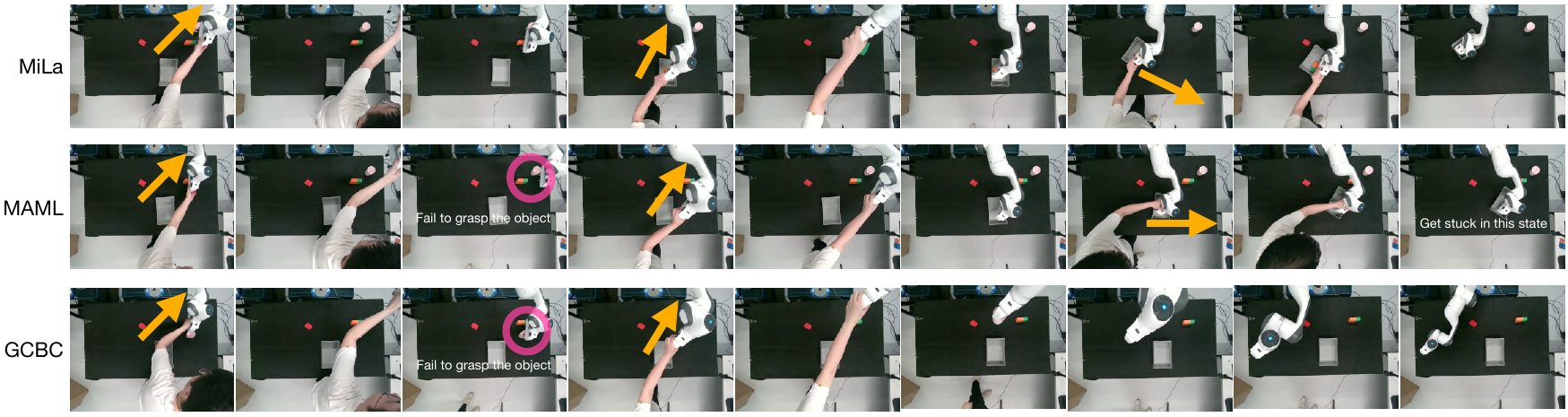}
    \caption{Snapshot of the long-horizon task in the presence of external disturbance and visual occlusion (i.e., human arm). \emph{First, second, and third row} correspond to evaluations using MiLa, MAML and GCBC, respectively.}
    \label{fig:snapshot-state-disturb}
\end{figure*}

In addition to visual occlusion, we consider exerting external perturbations on the robot arm directly. Specifically, the perturbations are imposed by dragging the robot's joints or end-effector arbitrarily.

In Fig.~\ref{fig:snapshot-state-disturb}, the yellow arrows depict the direction of perturbations. We can see that MiLa is the only method capable of recovering from the disturbances, whereas MAML and GCBC fail to do so, see the second and third rows of Fig.~\ref{fig:snapshot-state-disturb}. Note that MAML and GCBC may exhibit abnormal behaviours under perturbations. For example, at the last column of the second and third rows, MAML becomes immobilized and GCBC moves to an area that is far away from the demonstrated robot workspace.

The ability to resist disturbances exhibited by MiLa is attributed to the use of DMP to ensure goal-oriented motion, where the goal, inferred from visual observation at the start of each subtask, remains fixed throughout the subtask execution. In contrast, MAML and GCBC predict actions based on image observations and proprioceptive states at each time step. Any perturbations during task execution may disrupt these predictions, potentially resulting in task failure. We emphasize that external perturbations applied to the robotic arm also introduce visual disturbances, as the user's arm is unseen in the training dataset, evidencing that MiLa can resist disturbances in terms of visual observations and robot states.

\section{Conclusion} \label{sec:conclusion}

In this paper, we introduced a novel meta-imitation learning approach, MiLa, which is capable of rapidly learning new long-horizon visuomotor tasks and effectively resisting perturbations during task execution. MiLa establishes a skill repertoire capturing various elementary motion primitives and subsequently solving long-horizon tasks by reusing and composing motion primitives. Experimental results indicate that MiLa achieves superior performance compared to state-of-the-art baselines. Furthermore, our approach enables learning from unsegmented demonstrations and demonstrates robust resistance to perturbations from both visual inputs and the robot's proprioceptive states. 

While we assume that the order of subtasks is fixed, 
an important extension is to simultaneously learn both the order of primitives and their corresponding task parameters. 
As a long-term goal, it would be promising to extend MiLa to learn long-horizon tasks from cross-domain demonstrations, including those from humans and robots with different embodiments.

\bibliographystyle{unsrt}
\bibliography{bibiography}

\end{document}